\documentclass[conference]{IEEEtran}
\usepackage{times}
\usepackage{algorithm}
\usepackage{algorithmic}
\usepackage{amssymb}
\usepackage{amsmath}
\usepackage{amsfonts}
\usepackage{color}
\let\labelindent\relax
\usepackage{enumitem}
\usepackage{graphicx}
\usepackage{wrapfig}

\newcommand{\prg}[1]{\vspace{0.05in}\noindent\textbf{{#1}}}

\usepackage[numbers]{natbib}
\usepackage{multicol}
\usepackage[bookmarks=true]{hyperref}
\usepackage{color}

\begin{document}

\newcommand{\todo}[1]{\textcolor{red}{TODO: #1}}
\newcommand{\chelsea}[1]{\textcolor{magenta}{Chelsea: #1}}
\newcommand{\dorsa}[1]{\textcolor{blue}{Chelsea: #1}}

\title{Unsupervised Visuomotor Control through Distributional Planning Networks}


\author{\authorblockN{Tianhe Yu, Gleb Shevchuk, Dorsa Sadigh, Chelsea Finn}
\authorblockA{Stanford University\\
Email: \{tianheyu,glebs,dorsa,cbfinn\}@stanford.edu}
}


%

\maketitle

\begin{abstract}
While reinforcement learning (RL) has the potential to enable robots to autonomously acquire a wide range of skills, in practice, RL usually requires manual, per-task engineering of reward functions, especially in real world settings where aspects of the environment needed to compute progress are not directly accessible.
To enable robots to autonomously learn skills, we instead consider the problem of reinforcement learning without access to rewards.
We aim to learn an unsupervised embedding space under which the robot can measure progress towards a goal for itself.
Our approach explicitly optimizes for a metric space under which action sequences that reach a particular state are optimal when the goal is \emph{the final state reached.}
This enables learning effective and control-centric representations that lead to more autonomous reinforcement learning algorithms.
Our experiments on three simulated environments and two real-world manipulation problems show that our method can learn effective goal metrics from unlabeled interaction, and use the learned goal metrics for autonomous reinforcement learning. 
\end{abstract}

\IEEEpeerreviewmaketitle

\section{Introduction}

Reinforcement learning (RL) is a promising approach for enabling robots to autonomously learn a breadth of visuomotor skills such as grasping~\cite{pinto2016supersizing,kalashnikov2018qt}, object insertion and placement tasks~\cite{levine2016end}, and non-prehensile manipulation skills~\cite{dppt,dsae,chebotar2017combining}. However, reinforcement learning relies heavily on a reward function or metric that indicates progress towards the goal. In the case of vision-based skills, specifying such a metric is particularly difficult for a number of reasons. First, object poses are not readily accessible and pre-trained object detectors struggle without fine-tuning with data collected in the robot's domain~\cite{rosenfeld2018elephant}. Second, even when fine-tuned object detectors are available, the location of objects may not be a sufficient representation to identify success for some tasks, while a more suitable representation would require task-specific engineering. For example, if our goal is to manipulate a rope into a particular shape, the corresponding reward function would need to detect the shape of the rope. In many ways, such task-specific engineering of rewards defeats the point of autonomous reinforcement learning in the first place, as the ultimate goal of RL is to eliminate such manual and task-specific efforts.


Motivated by this problem, one appealing alternative approach is to provide an example image of a desired goal state~\cite{deguchi1999image,e2c,dsae,upn,imaged_goal,edwards2017perceptual}, and derive a reward function using the goal image. While such goal observations are applicable to a variety of goal-centric tasks and often easy for a user to provide, they do not solve the problem of rewards entirely: na\"{\i}ve distances to the goal image, such as mean squared error in pixel space, do not provide a suitable metric space for reinforcement learning as they are sensitive to small changes in lighting, differences in camera exposure, and distractor objects. 
In this paper, our goal is to leverage \emph{autonomous, unlabeled interaction data} to learn an underlying informative metric that can enable the robot to achieve a variety of goals with access to only a single image of the task goal.
This capability would enable reinforcement learning of such tasks to be significantly more autonomous.

\begin{figure}[!t]
    \centering
    \includegraphics[width=\columnwidth]{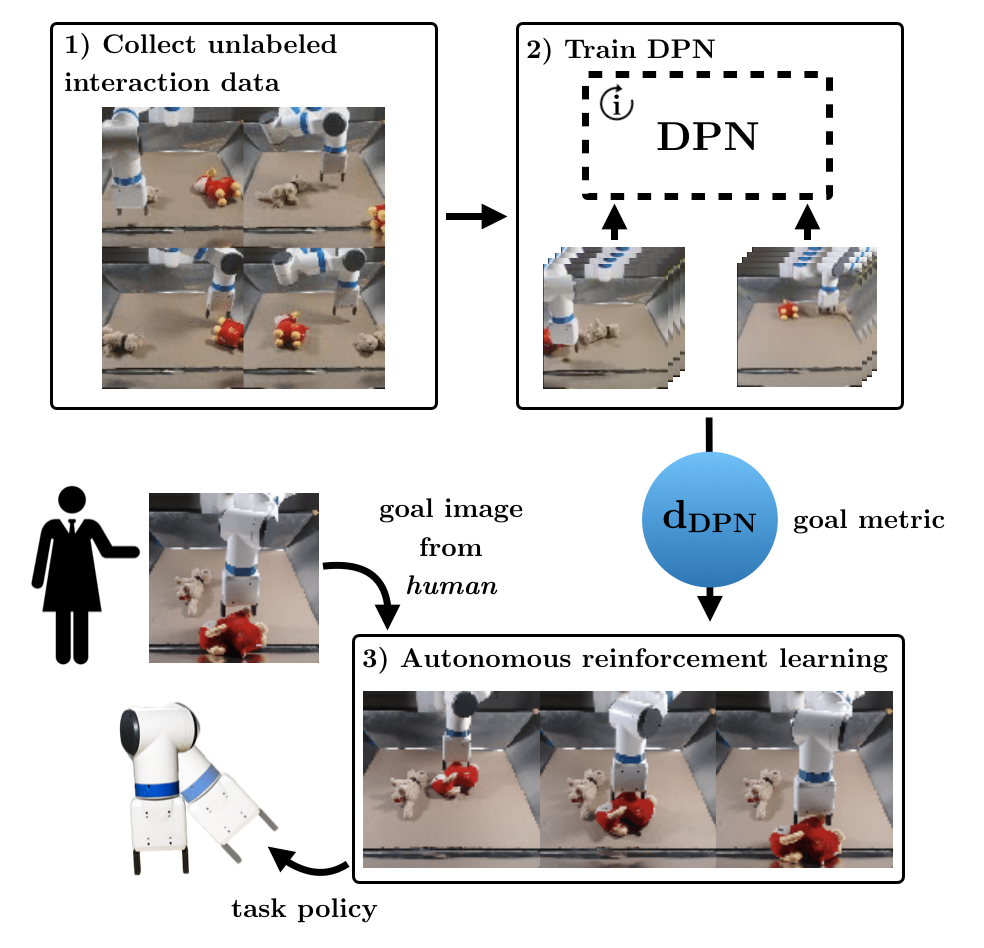}
    \vspace{-0.3cm}
    \caption{General overview of our method. Our method, DPN, enables autonomous reinforcement learning, without human-provided reward functions, on vision-based manipulation problems.
    }
    \vspace{-0.3cm}
    \label{fig:teaser}
\end{figure}

To approach this problem, we aim to learn an embedding space that imposes a metric with respect to a goal image, without using human supervision.
One natural option is to use unsupervised representation learning methods~\cite{e2c,dsae,imaged_goal}. However, these models are largely trained as density estimators, meaning that they will pay attention to the most salient aspects of the images rather than the ones that are relevant for control.
Instead, our goal is to learn a \emph{control-centric} representation that takes into account how a sequence of actions leads to a particular observation and ignores other changes in the observation space that are not caused by actions. 

Our key insight is that any sequence of actions is optimal under the binary reward function of reaching the final state resulting from those actions. 
Further, we can use this property to explicitly optimize for control-centric metric spaces from unsupervised interactions. 
In particular, we propose to explicitly optimize for a metric such that the sequences of actions that lead to a given goal image have high-likelihood when optimizing with respect to the metric. 
Our approach can be viewed as a generalization of universal planning networks~\cite{upn} to distributions of actions, while critically showing that such models can be trained from real-world unsupervised interaction rather than simluated expert demonstration data. 
Our experiments on three simulated domains and two real-world domains demonstrate that our approach can effectively enable robots to learn reaching, object pushing, and rope manipulation tasks from raw pixel observations without human reward feedback and with minimal engineering.

\section{Related Work}

Our work aims to enable a robot to learn a variety of skills without human supervision, hence falling under the category of self-supervised robotic learning~\cite{pinto2016supersizing,agrawal2016learning,Finn2017DeepVF,byravan2017se3,levine2018learning}. We specifically approach this problem from the perspective of  representation learning, using the learned embedding as a goal metric for reinforcement learning for reaching goal images. Prior works have aimed to learn representations for control through auto-encoding~\cite{lange2012autonomous,e2c,dsae,dppt,imaged_goal}, pre-trained supervised features~\cite{sermanet2016unsupervised}, spatial structure~\cite{dsae,dppt,jonschkowski2017pves}, and viewpoint invariance~\cite{pierre_time}. However, unlike these works, we build a metric that specifically takes into account how actions lead to particular states, leading to control-centric representations that capture aspects of the observation that can be controlled, while discarding other elements.

Previous approaches to building control-centric goal representations include using inverse models~\cite{agrawal2016learning,pathak2018zero} and mutual information estimation~\cite{discern}. Unlike our approach, these methods will not necessarily encode all of the aspects of an observation needed to reach a goal. Further, inverse models are susceptible to local optimum when planning greedily. Other methods have built effective reward functions or goal metric spaces using either expert demonstration data~\cite{Abbeel04,Ziebart08,gcl,upn}, pre-trained goal-conditioned policies~\cite{ghosh2018learning}, or other forms of supervision~\cite{Daniel-RSS-14,edwards2017cross}. Our approach, on the other hand, does not use supervision and requires only unlabeled interaction data.

Related to learning \emph{goal} representations, a number of prior works have considered the problem of learning control-centric \emph{state} representations from interaction data~\cite{rosencrantz2004learning,boots2011closing,jonschkowski2015learning,lesort2018state,thomas2017independently,lesort2017unsupervised}, for use with planning or reinforcement learning under a known reward or cost.
Other works have combined auxiliary representation learning objectives with reinforcement learning~\cite{de2018integrating,jaderberg2016reinforcement,shelhamer2016loss}. Unlike all of these methods, we focus on representations that induce accurate and informative goal metrics and do not assume access to any reward functions or metrics on state observations.



\newcommand{\loss}{\mathcal{L}}
\newcommand{\action}{\mathbf{a}}
\newcommand{\obs}{\mathbf{o}}
\newcommand{\z}{\mathbf{z}}
\newcommand{\x}{\mathbf{x}}
\newcommand{\state}{\mathbf{x}}

\newcommand{\enc}{f} 
\newcommand{\encparams}{\theta_\text{enc}}
\newcommand{\dyn}{g} 
\newcommand{\dynparams}{\theta_\text{dyn}}
\newcommand{\dec}{h} 
\newcommand{\decparams}{\theta_\text{act}}

\section{Preliminaries}
\label{sec:prelim}
\vspace{-0.1cm}

Our method builds upon universal planning networks (UPN)~\cite{upn}, which learn abstract representations for visuomotor control tasks using expert demonstration trajectories. The representation learned from UPN provides an effective metric that can be used as a reward function to specify new goal state from images in model-free reinforcement learning.

To learn such representations, the UPN is constructed as a model-based planner that performs gradient-based planning in a latent space using a learned forward dynamics model. UPN encodes the initial image of a particular control task into the latent space, and then iteratively executes plans to reach the latent representation of the goal image. The plans are selected via gradient descent on the latent distance between the predicted terminal state and the encoding of the actual target image. Simultaneously, an outer imitation objective ensures that the learned plans match the sequences of actions of the training demonstrations. Consequentially, UPNs learn a latent distance metric by directly optimizing a plannable representation with which gradient-based planning leads to the desired actions.

\begin{figure*}[!t]
    \centering
    \includegraphics[width=0.8\linewidth]{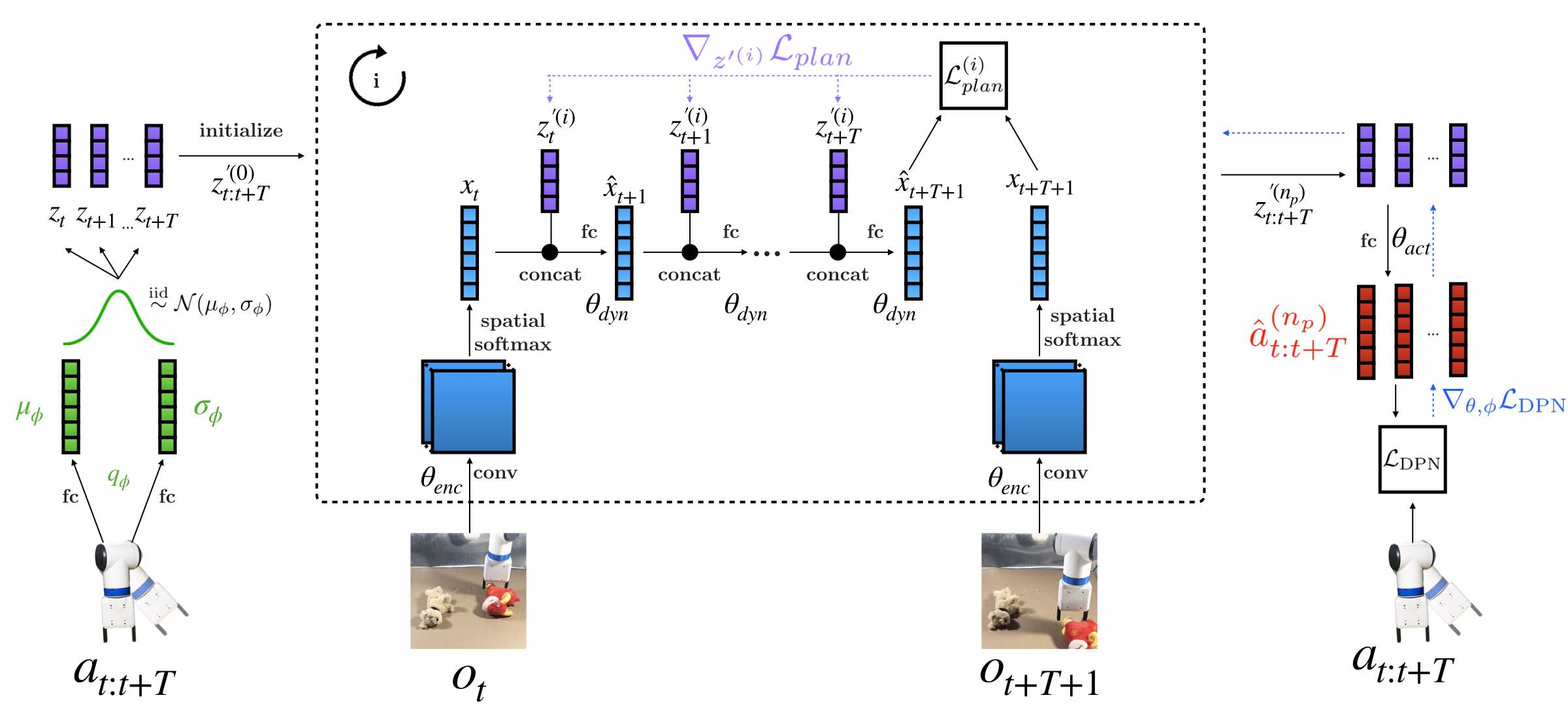}
    \vspace{-0.4cm}
    \caption{Diagram of our distributional planning networks model. Our model enables learning a representation $\x$ that induces a control-centric goal metric on images $\obs$ from unlabeled interaction data. It does so by explicitly training for a metric under which gradient-based planning leads to the a sequence of actions that reach the final image. To effectively model the many action sequences that might lead to a goal after $T$ timesteps, we introduce latent variables $\z_{t:t+T}$ and train the model using amortized variational inference.
    }
    \vspace{-0.5cm}
    \label{fig:diagram}
\end{figure*}

Concretely, given initial and goal observations $\obs_t$ and $\obs_g$, e.g., as seen in the two images in Fig.~\ref{fig:diagram}, the model uses an encoder $\enc$ to encode the images into latent embeddings: 
$$\x_{t} = \enc(\obs_t; \encparams) \:\:\:\:\:\:\: \x_g = \enc(\obs_g; \encparams),$$ 
where $\enc(\cdot, \encparams)$ is a convolutional neural network. After encoding, the features $\x_t$ and $\x_g$ are fed into a gradient descent planner (GDP), which outputs a predicted plan $\hat{\action}_{t:t+T}$ to reach $\x_g$ from $\x_t$. The GDP is composed of a forward dynamics model $\dyn$ with parameters $\dynparams$ where $\hat{\x}_{t+1} = \dyn(\x_t, \hat{\action}_t; \dynparams)$. The learned plan is initialized randomly from a uniform distribution $\hat{\action}^{(0)}_{t:t+T} \sim \mathcal{U}(-1, 1)$ and is updated iteratively via gradient descent as follows: $$\hat{\action}^{(i+1)}_{t:t+T} = \hat{\action}^{(i)}_{t:t+T} - \alpha\nabla_{\hat{\action}^{(i)}_{t:t+T}}\loss^{i}_{plan},$$ 
where $\alpha$ is the gradient descent step size and $\loss^{i}_{plan} = \|\hat{\x}^{(i)}_{t+T+1} - \x_g\|_2^2$. In practice, we find the Huber loss is more effective than the $\ell_2$ loss for $\loss_{plan}$. After computing the predicted plan, UPN updates the planner by imitating the expert actions $\action^*_{t:t+T}$ in the outer loop. The imitation objective is computed as $\loss_{imitation} = \|\hat{\action}_{t:t+T} - \action^*_{t:t+T}\|_2^2$ and is used to update the parameters of the encoder and forward dynamics $\theta := \{\encparams, \dynparams\}$ respectively:
$$\theta \leftarrow \theta - \beta\nabla_{\theta}\loss_{imitation} 
$$ 
where $\beta$ is the step size for the outer gradient update.

\citet{upn} applied the learned latent metric as a reward function for model-free reinforcement learning to a range of visuomotor control tasks in simulation and showed that the robot can quickly solve new tasks with image-based goals using the latent metric. However, in order to learn effective representations for new tasks, UPNs require access to optimal expert demonstrations, which are difficult and time-consuming to collect, making it difficult to extend to a variety of tasks, in particular, real-world tasks. In response, we will show a key extension to UPNs that can effectively learn such latent representations without using expert demonstrations in the next section.

\section{Unsupervised Distributional Planning Networks}

Our end goal is to enable a robot to use reinforcement learning to reach provided goal images, without requiring manually-provided or hand-engineered rewards. To do so, we will derive an approach for learning a metric space on image observations using only unsupervised interaction data. Universal planning networks (UPNs)~\cite{upn} show how we can learn such a metric from demonstration data. Further, \citet{ghosh2018learning} observe that one can learn such a metric with access to a goal-conditioned policy by optimizing for a goal metric that reflects the number of actions needed to reach a particular goal. However, if our end-goal is to learn a policy, we are faced with a ``chicken-and-egg'' problem: \emph{does the goal-conditioned policy come first or the goal metric?} To solve this problem, we propose to learn both at the same time. Our key observation is that a sequence of any actions, even random actions, is optimal under the binary reward function of reaching the final state resulting from those actions. Specifically, we can use random interaction data to optimize for a metric such that the following is true: when we find a sequence of actions that minimizes the distance metric between the final predicted embedding and the embedded goal image, the true sequence of actions has high probability. Concretely, consider interaction data consisting of an initial image $\obs_1$, a sequence of actions $\action_{1:t-1}$ executed by the robot, and the resulting image observation $\obs_t$. We can use data like this to optimize for a latent space $\x = \enc(\obs; \encparams)$ such that when we plan to reach $\x_t = \enc(\obs_t; \encparams)$, we have high probability of recovering the actions taken to get there, $\action_{1:t-1}$.

So far, this computation is precisely the same as the original UPN optimization, except that we perform the optimization over randomly sampled interaction data, rather than demonstrations. In particular, we surpass the need for expert demonstrations because of the observation that random interaction data can also be viewed as ``expert'' behavior with respect to the cost function of reaching a particular goal observation at the last timestep of a trajectory (whereas the original UPN was optimizing with respect to the cost function of reaching a goal state with a sequence of optimally short actions). Once we have a representation that can effectively measure how close an observation is to a goal observation, we can use it as an objective that allows us to optimize for reaching a goal observation quickly and efficiently, even though the data that was used to train the network did not reach goals quickly.
While not all robotic tasks can be represented as reaching a particular goal observation, goal reaching is general to a wide range of robotic control tasks, including object arrangement such as setting a table, deformable object manipulation such as folding a towel, and goal-driven navigation, such as navigating to the kitchen.

However, note that, unlike in the case of expert demonstration data, we are no longer optimizing for a unique solution for the sequence of actions: there are multiple sequences of actions that lead to the same goal. Hence, we need to model all of these possibilities.  We do so by modeling the distribution of action sequences that could achieve the goal, in turn training the UPN as a stochastic neural network to sample different action sequences.
Interestingly, universal planning networks are already stochastic neural networks, since the initial action sequence is randomly sampled before each planning optimization. However, as described in Section~\ref{sec:prelim}, they are trained with a mean-squared error objective, which encourages the model to represent uncertainty by averaging over possible outcomes. To more effectively model the multiple possible sequences of actions that can lead to a potential goal observation, we extend universal planning networks by enabling them to sample from the distribution of potential action sequences. To do so, we introduce latent variables into the UPN model and build upon advances in amortized variational inference~\cite{vae,johnson2016composing} to train the model, which we will discuss next.

\subsection{Distributional Planning Network Model}
To extend universal planning networks towards modeling distributions over actions, we introduce latent variables into the model. 
We thus consider the following distribution,
\begin{align*}
    &p(\action_{t:t+T} | \obs_t, \obs_{t+T+1}) \\ &=
    \int p(\action_{t:t+T} | \z_{t:t+T}, \obs_t, \obs_{t+T+1}) p(\z_{t:t+T}) d\z_{t:t+T},
\end{align*}
by introducing latent variables $\z_t$ for each timestep $t$. We model the prior over each timestep independently, and model each marginal as a standard Gaussian:
\begin{align*}
p(\z_{t:t+T}) = \prod_{t'=t}^{t+T} p(\z_{t'})  ~~~~~~~~~~~ p(\z_{t}) = \mathcal{N}(\mathbf{0},I) 
\end{align*}
We model $p(\action_{t:t+T} | \z_{t:t+T}, \obs_t, \obs_{t+T+1})$ using a neural network with parameters $\theta$ with two components. The first component is a deterministic gradient descent action planner with respect to latent action representations $\z'_{t:t+T}$, with gradient descent initialized at $\z_{t:t+T}$.
The second component is a feedforward decoder that maps from an individual latent action representation $\z'_t$ to a probability distribution over the corresponding action $\action_t$. We will next describe these two components in more detail before discussing how to train this model.

\noindent Concretely, the gradient-based planner component consists of:
\begin{description}
    \item[(a)] an encoder $\enc(\cdot; \encparams)$, 
which encodes the current and goal observation $\obs_t, \obs_g$ into 
the latent state space $\x_t, \x_g$, 
 \item[(b)] a latent dynamics model $\hat{\x}_{t+1} = \dyn(\x_{t}, \z'_t; \dynparams)$ that now operates on latent actions $\z'_t$ rather than actions $\action_t$, and
 \item[(c)] a gradient descent operator on $\z'_{t:t+T}$ that is initialized at a sample from the prior $\z'^{(0)}_{t:t+T}=\z_{t:t+T}$, and runs $n_p$ steps of gradient descent to produce $\z'^{(n_p)}_{t:t+T}$, using learned step size $\alpha_i$ for step $i=1,...,n_p$.
\end{description}
Like before, the gradient descent operator computes gradients with respect to the planning loss $\loss_{plan}$, which corresponds to the Huber loss between the predicted $\hat{\x}_{t+T+1}$ and the encoded goal observation, $\x_g$.

Once we have computed a sequence of latent actions $\z'^{(n_p)}_{t:t+T}$ using the planner, we need to decode the latent values into actions. We do so using a learned action decoder $\dec(\action_t | \z'_t; \decparams)$. This feedforward neural network outputs the mean of a Gaussian distribution over the action with a fixed constant variance. Overall, the parameters of our model are $\theta = \{\encparams, \dynparams, \decparams, \alpha_i\}$. The architectures for each of these components are described in detail in Appendix~\ref{app:details}. We next describe how to train this model.

\subsection{Distributional Planning Network Training}
\newcommand{\dpnloss}{\loss_\text{DPN}}

Since we are training on random interaction data, there are many different sequences of actions that may bring the robot from one observation to another. To effectively learn from this data, we need to be able to model the distribution over such action sequences. To do so, we train the above model using tools from amortized variational inference. We use an inference network to model the the variational distribution, which is factorized as 
$$q_\phi(\z_{t:t+T}| \action_{t:t+T}) = \prod_{t'=t}^{t+T} q(\z_{t'} | \action_{t'}; \phi),
$$
where $q(\z_{t} | \action_{t}; \phi)$ outputs the parameters of a conditional Gaussian distribution $\mathcal{N}(\mu_\phi(\action_{t}), \sigma_\phi(\action_t))$.
Following~\citet{vae}, we use this estimated posterior to optimize the variational lower bound on the likelihood of the data:
\begin{align}
    \dpnloss(\theta, \phi) = &- \mathbb{E}_{\z_{t:t+T} \sim q_\phi} \nonumber  
    \left[ \log p_\theta (\action_{t:t+T} | \obs_t, \obs_{t+T+1})    \right]\\
    &+ \beta D_\text{KL}( q_\phi(\z_{t:t+T}| \action_{t:t+T}) ~|| ~ p(\z_{t:t+T})).
\end{align}
A value of $\beta=1$ corresponds to the correct lower bound. As found in a number of prior works (e.g.~\cite{betavae,babaeizadeh2017stochastic}), we find that using a smaller value of $\beta$ leads to better performance.
We compute this objective using random interaction data that is autonomously collected by the robot, and optimize it with respect to the model parameters $\theta$ and the inference network parameters $\phi$ using stochastic gradient descent. 
Mini-batches are sampled by sampling a trajectories from the dataset, $\obs_1, \action_1, ...$, and selecting a length-$T$ segment at random within that trajectory: $\obs_t, \action_t, ..., \action_{t+T}, \obs_{t+T+1}$.
We compute the objective using these sampled trajectory segments by first passing the executed action sequence into the inference network to produce a distribution over $\z_{t:t+T}$. 
The second term in the objective operates on this Gaussian distribution directly, while the first term is computed using samples from this distribution. In particular, we compute the first term by passing observations $\obs_t, \obs_{t+T+1}$ and the samples $\z_{t:t+T}$ as input to the gradient descent planner, running gradient descent for $n_p$ timesteps, and decoding the result into $\hat{\action}_{t:t+T}$ to produce the distribution $p_\theta (\action_{t:t+T} | \obs_t, \obs_{t+T+1}) = \mathcal{N}(\hat{\action}_{t:t+T}, I)$. See Figure~\ref{fig:diagram} for a summary of the model and training.

\subsection{RL with the Learned Goal Metric}


Training the distributional planning network provides us with several key components. Most importantly, the encoder $\enc(\cdot; \encparams)$ of the DPN provides an embedding of images under which distances to goal images accurately reflect whether or not a sequence of actions actually reached the goal.
The combination of the image encoder $\enc$, latent dynamics $\dyn$, and action decoder $\dec$ serves as a policy that can optimize over a sequence of actions that will reach an inputted goal image.


One easy way to use the DPN model is directly as a goal-conditioned policy. In particular, consider a human-provided goal image $\obs_g$ for a desired task. We can compute a sequence of actions to reach this goal image $\hat{\action}_{t:t+T}$ by running a forward pass through the DPN with $\obs_t$ and $\obs_g$ as input and gradient descent initialized at a sample from the prior $\z'^{(0)}_{t:t+T} \sim p(\z_{t:t+T})$.
However, note that the model outputs a distribution over \emph{all} action sequences that might reach the final state after $T$ actions. This means that we can expect the action sequence produced by DPN to reach the goal, but may not do so in a timely manner. In turn, the DPN encoder represents a true metric of whether or not an embedded image $\x$ has reached the same state as another embedded image $\x'$, as it is minimizes for an action sequence that reaches the correct image. 
As a result, we can alternatively use this metric space with reinforcement learning to optimize for efficiently reaching a goal image.

Thus, after training the DPN model on autonomous, unlabeled interaction, we discard most of the DPN model, only keeping the encoder $\enc(\cdot; \encparams)$: this encoder provides a goal metric on images. To enable a robot to autonomously learn new tasks specified by a human, we assume a human can provide an image of the goal $\obs_g$, from the perspective of the robot.
We then run reinforcement learning, without hand-tuned reward functions, by deriving a reward function from this goal metric. We derive rewards according to the following equation:
$$
r(\obs_t; \obs_g) = - \exp( \loss_\delta(\obs_t, \obs_g) )
$$
where $\loss_\delta$ corresponds to the Huber loss:
$$\loss_\delta(\obs_t, \obs_g) = \|d_\text{DPN}(\enc(\obs_t;\encparams) - \enc(\obs_g;\encparams), \delta)\|_1$$
where for the $i$-th entry $\x_i$ of some vector $\x$,
$$
d_\text{DPN}(\x, \delta)_i \!=\! \begin{cases}
 \frac{1}{2}\x_i^2     & \!\!\textrm{for } |\x_i|  \le \delta, \\
 \delta\, |\x_i| - \frac{1}{2}\delta^2 & \!\!\textrm{otherwise.}
\end{cases}
$$
Following~\citet{upn}, we use $\delta=0.85$. We then use the soft actor-critic (SAC) algorithm~\cite{sac} for running reinforcement learning with respect to this reward function.

\section{Experiments}
\label{sec:experiments}
The goal of our experiments is to test our primary hypothesis: \emph{can DPN learn an accurate and informative goal metric using only unlabeled experience?} We design several simulated and real world experiments in order to test this hypothesis with both synthetic and real images in multiple domains, ranging from simple visual reaching tasks to more complex object arrangement problems. In all cases, the objective is to reach the goal state which is illustrated to the agent by a goal image. 
We will release our code upon publication, and you can find videos of all results at the following link\footnote{The supplementary website is at \url{https://sites.google.com/view/dpn-public}}.

To quantify the performance of DPN, we compare our method to leading prior approaches for learning goal metrics from unlabeled interaction data. In particular, we compare to the following approaches:
\begin{itemize}
    \item We train a multi-step \textbf{inverse model} to predict the intermediate actions $\action_{t:t+T}$ given two observations $\obs_t, \obs_{t+T+1}$. Following~\citet{agrawal2016learning} and~\citet{pathak2018zero}, we use a siamese neural network that first embeds the two observations and then predicts the actions from the concatenated embeddings. We include a forward-consistency loss as a regularizer of the inverse model suggested in~\citet{pathak2018zero}. We use the embedding space as a goal metric space.
    \item We train a variational autoencoder (\textbf{VAE})~\cite{vae}, and use distances in the latent space as a goal metric, as done by prior work~\cite{imaged_goal}.
    \item We lastly evaluate $\ell_2$ distance in \textbf{pixel space} as the goal metric.
\end{itemize}
All of the above approaches are trained on the same unlabeled datasets for each problem domain, except for the pixel distance metric, which is not learned. Because inverse models have a tendency to only pay attention to the most prominent features of an image to identify the actions, we expect the inverse models to work best in situations where the robot's embodiment is a critical aspect of the goal, such as reaching tasks, rather than tasks involving objects. However, even in situations such as reaching, the metric underlying the learned embedding may not correspond to true distances between states in a meaningful way.
On the other hand, because VAEs learn an embedding space by reconstructing the input, we expect VAEs to work best in situations where the goal involves particularly salient parts of the image. Finally, we do not expect pixel error to work well, as matching pixels exactly is susceptible to local minima and sensitive to lighting variation and sharp textures. 




Finally, we use Soft Actor-Critic (SAC)~\cite{sac} with default hyperparameters as the RL algorithm for training all experiments with all four metrics respectively.

We provide full hyperparameters, architecture information, and experimental setup details in Appendix~\ref{app:details} in the supplemental material.

\begin{figure}[!t]
    \centering
    \includegraphics[width=\linewidth]{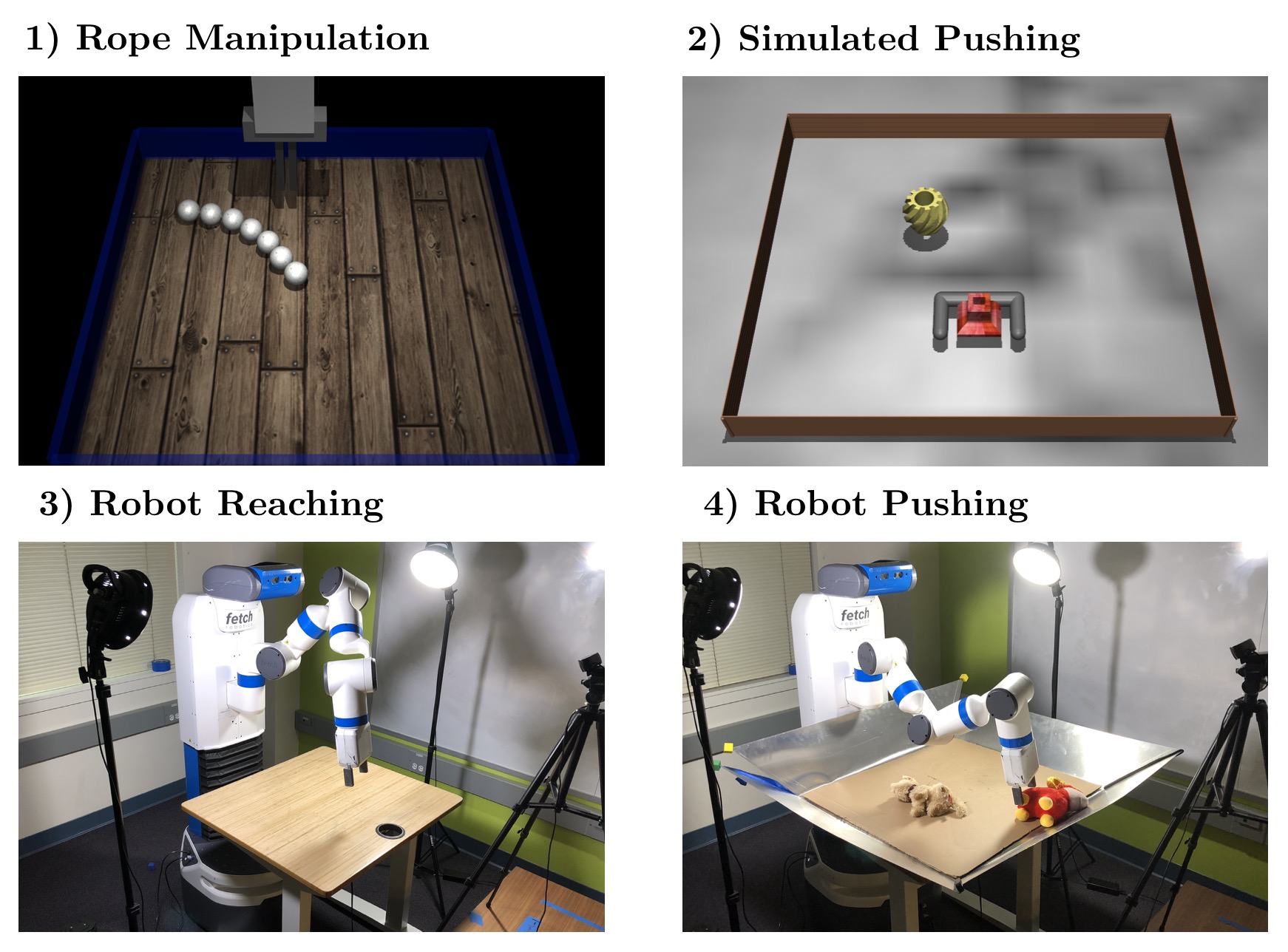}
    \vspace{-0.3cm}
    \caption{We conduct experiments on several different vision-based manipluation domains, including simluated rope manipulation, simulated pushing, robot reaching, and robot pushing in the real world.
    }
    \vspace{-0.3cm}
    \label{fig:setup}
\end{figure}

\subsection{Simulation Experiments}

We evaluate our approach starting from simulated experiments using the MuJoCo physics engine~\cite{mujoco}. For all simulated experiments, the inputs $\obs_t$ and $\obs_g$ are $100 \times 100$ RGB images.
\begin{figure*}[!t]
    \centering
    \includegraphics[width=0.3\linewidth]{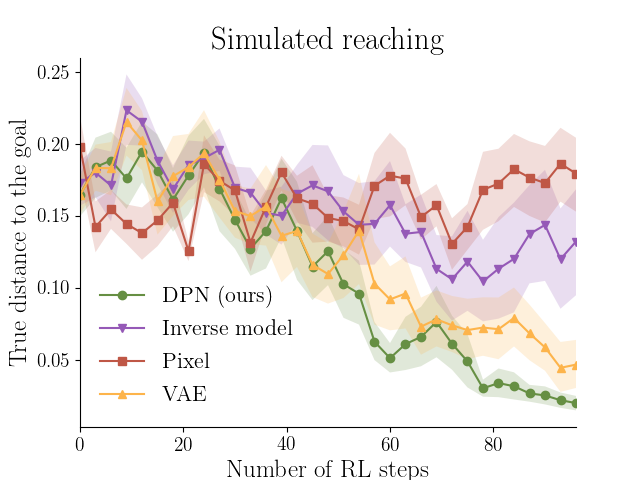}
    \includegraphics[width=0.3\linewidth]{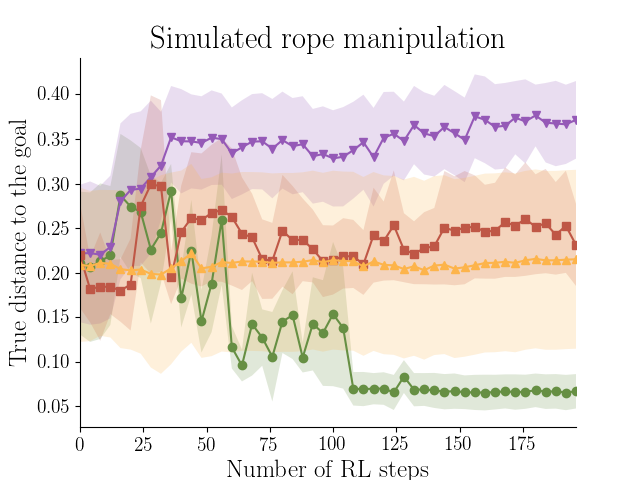}
    \includegraphics[width=0.3\linewidth]{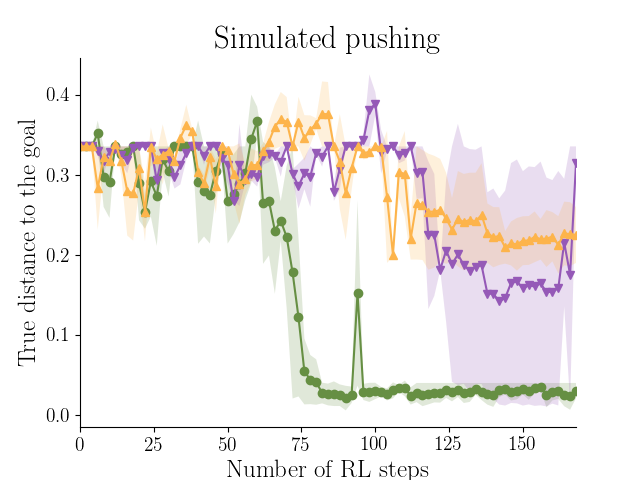}
    \vspace{-0.2cm}
    \caption{Quantitative simulation results that evaluate the effectiveness of the goal metrics induced by each method by measuring the true distance to the goal state when running reinforcement learning with the reward derived from the learned goal metric. Performance is averaged across multiple tasks and error bars indicate standard error. Each RL step requires $20$ samples from the environment.
    }
    \vspace{-0.2cm}
    \label{fig:sim_results}
\end{figure*}

\begin{figure*}[!t]
    \centering
    \includegraphics[width=0.9\linewidth]{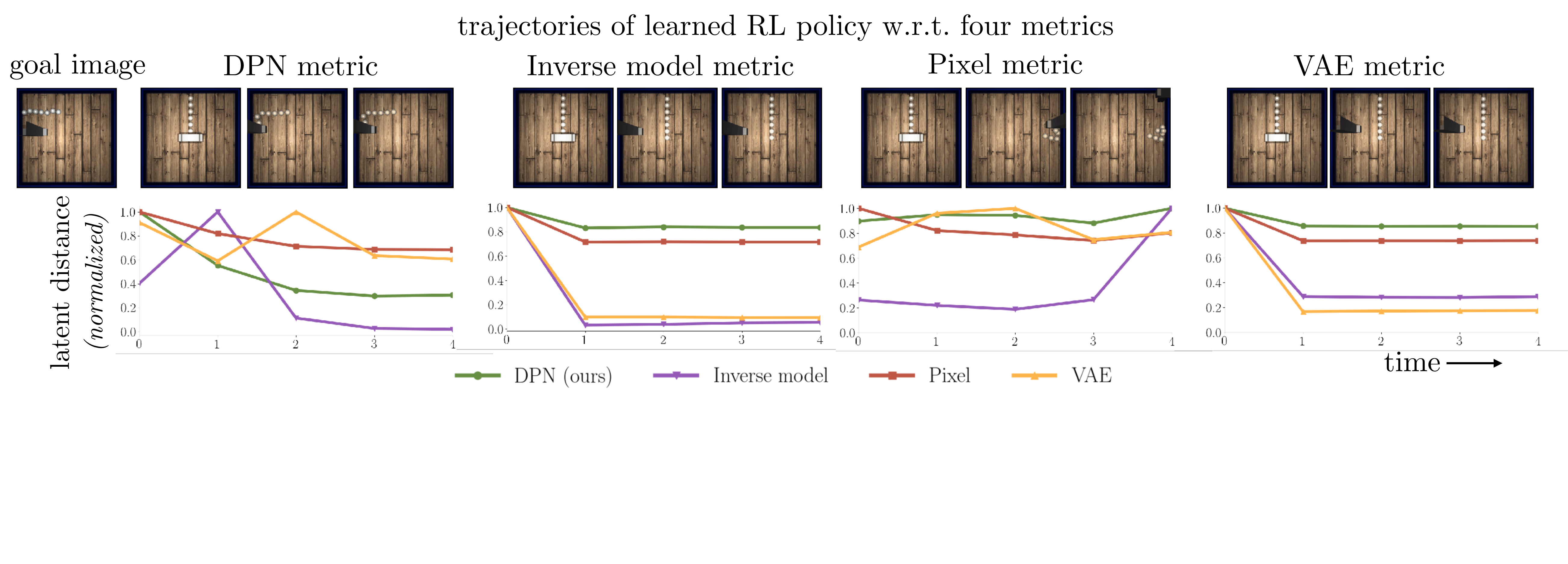}
    \vspace{-2cm}
    \caption{Comparisons of normalized latent distance to the goal determined by four approaches for the simulated rope manipulation task. We evaluate each latent metric on the trajectories (from a top-down view) of RL policy with respect to DPN, inverse model, VAE, and pixel space, shown above from left to right. Note in the leftmost plot that, though the metric learned by the inverse model achieves a lower normalized latent distance than the DPN metric, it goes to around $0$ once the gripper moves closer to its corresponding position in the goal image without touching the rope as shown in the second and fourth plot from the left. This suggests that the inverse model metric fails to capture the actual goal of task, which is directing the rope to the right form.
    }
    \vspace{-0.2cm}
    \label{fig:rope_metric}
\end{figure*}

\begin{figure*}[!t]
    \centering
    \includegraphics[width=0.8\linewidth]{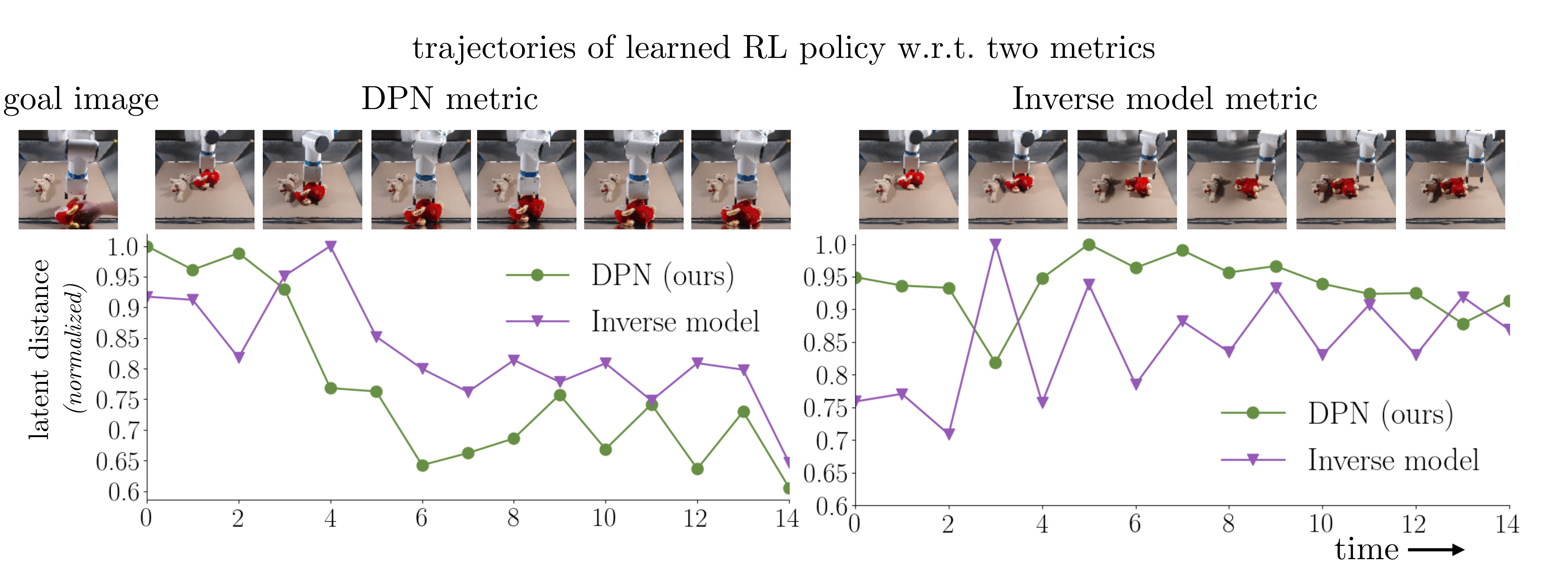}
    \vspace{-0.3cm}
    \caption{Comparisons of normalized latent distance to the goal determined by DPN and inverse model for the real-world pushing task. We evaluate each latent metric on the trajectories of RL policy with respect to DPN and inverse model respectively, shown in the images above from left to right.
    }
    \vspace{-0.2cm}
    \label{fig:fetch_push_metric}
\end{figure*}

\begin{figure}[!t]
    \centering
    \includegraphics[width=\linewidth]{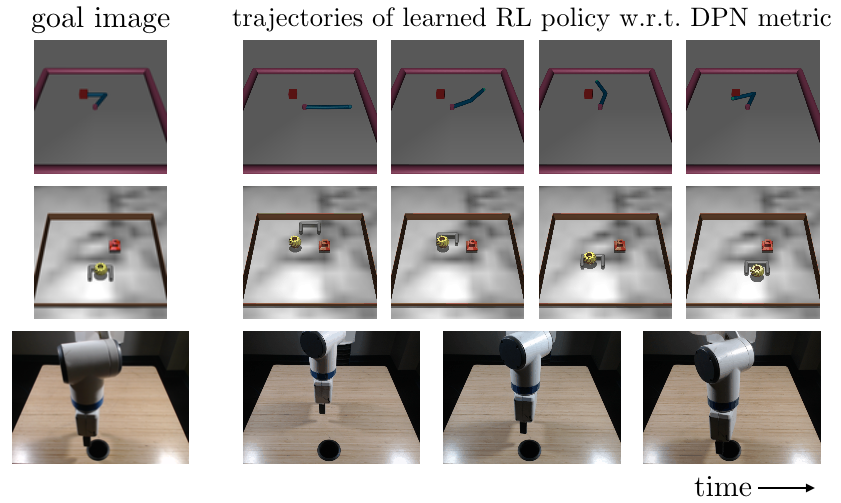}
    \vspace{-0.4cm}
    \caption{Roll-outs of learned RL policy using the DPN metric of simulated reaching, simulated pushing, and robot reaching experiments from top to bottom.
    }
    \vspace{-0.5cm}
    \label{fig:film_strip}
\end{figure}

\prg{Simulated Reaching.}
The first experimental domain is a planar reaching task, where the goal of the task is for a $2$-link arm to reach a colored block. 
We collect $30000$ videos of unlabeled physical interactions and train DPN, the inverse model, and the VAE on the random dataset. Note that since this is a fully observed setup with no object interaction, we do not use the multi-step inverse model with recurrence and instead use a one-step inverse model with feed-forward networks, as suggested by~\citet{pathak2018zero}.

We evaluate all learned metrics along with the pixel space metric by running SAC on $10$ different tasks where the target block is at a different position for each task. The input to the RL policy is the joint angles of the robot. We summarize the comparison in Figure~\ref{fig:sim_results}. As shown in the plot, our method is able to reach the goal within $0.05$cm in $60$ RL steps and gets closer to $0$ after 100 steps (see Figure~\ref{fig:film_strip}). The RL policies with metrics learned by VAE and the inverse model are also able to get to the proximity of the goal but are less accurate. This is reasonable since VAEs usually pay attention to the most salient part of the image while inverse models usually pay attention to objects that correlate most with the actions, i.e. the robot arm in this domain. Tracking the arm movement is sufficient to solving reaching tasks. The pixel space distance, meanwhile, struggles to find the goal as expected since pixel-wise distance is susceptible to minor environmental changes.

\prg{Simulated Rope Manipulation.}
The goal of the second experiment is to manipulate a rope of $7$ pearls into various shapes using a parallel-jaw gripper, where the setup is shown in Figure~\ref{fig:setup}. In this experiment, we aim to test if our method can focus on the shape of the rope rather than the position of the gripper since, unlike simulated reaching, only encoding the movement of the gripper into the latent space would lead to ignoring the actual goal of the task: manipulating the rope.

We collect $20000$ 10-frame random videos. Similar to simulated reaching, we then train a one-step inverse model for rope manipulation.

We evaluate the four metrics by running SAC on $4$ tasks where the rope is displaced to a different shape in each task. The input to the RL policy are the end-effector positions and velocities of the gripper. For evaluation, we define the true distance to goal following~\citet{xie2018fewshotgoal}, measuring the average distance of corresponding pearls in the rope. As seen from the results in Figure~\ref{fig:sim_results}, our method does substantially better than the other three approaches, achieving around $0.05$cm on average to the shapes shown in the goal images. The other three approaches fail to lead to effective RL training. To conduct a more direct comparison of all the latent metrics, we plot the latent distance to the goal of four approaches when rolling out the trajectories of learned RL policy with all the four metrics as reward functions respectively in Figure~\ref{fig:rope_metric}. Notice that the metrics learned by the inverse model and the VAE go to around $0$ once the gripper goes to its corresponding position in the goal image but completely disregards the rope. In contrast, the DPN metric only decreases when the rope is manipulated to the target shape. This experiment demonstrates that DPN is able to produce a more informative metric without collapsing to the most salient feature in the image.

\prg{Simulated Pushing.} 
For our third simulated experiment, we perform a simulated pushing task where a robot arm must push a target object to a particular goal position amid one distractor. In order to make this environment more realistic, we use meshes of a vase and a strainer from \url{thingiverse.com} with different textures for the two objects and a marble texture for the table (see Figure~\ref{fig:setup}).

For this task, we collect $3000$ random $16$-frame videos and train a multi-step recurrent inverse model for comparison.

Based on the previous two experiments, the pixel distance does not serve as a good metric, so we drop it and only evaluate DPN, inverse models and VAEs. For this task, knowing only the hand's position is not sufficient to solve the task. Hence, in addition to end-effector positions and velocities of the robot hand, a latent representation of the current image extracted from each method respectively is also provided as an input the RL policy. As seen in Figure~\ref{fig:sim_results}, DPN learns to push both objects toward the goal position with a distance close to $0$cm (see Figure~\ref{fig:film_strip} for an example). The multi-step inverse model is only able to push one of the objects to goal, as indicated by the large standard error in Figure~\ref{fig:sim_results}. The VAE metric can not quite learn how to push both objects and RL training does not make significant progress.

\subsection{Real World Robot Experiments}

In order to ascertain how well our approach works in the real world with real images, we evaluate on two robot domains. Similar to the simulated tasks, a robot, a Fetch Manipulator, needs to reach a goal image using images from a forward-facing camera by applying continuous end effector velocities. Both setups are shown in Figure~\ref{fig:setup}.

\prg{Robot Reaching.}
First, we evaluate our method for a simple environment, where the robot must learn to move its end-effector along the xy plane to a certain position above a table.
We collect unlabeled interaction data by having the arm randomly move above the table. For this task, we capture $5000$ episodes, which corresponds to approximately 28 hours of capture.

We train DPN on $4000$ samples and use the remaining $1000$ samples for validation. Seeing that the inverse and VAE models are the next best-performing in simulation experiments, we also train both on this dataset using the same procedures described earlier. We also use the one-step inverse model for this experiment with the same reason discussed in the simulated reaching section.

\begin{figure} 
    \centering
    \includegraphics[width=0.85\linewidth]{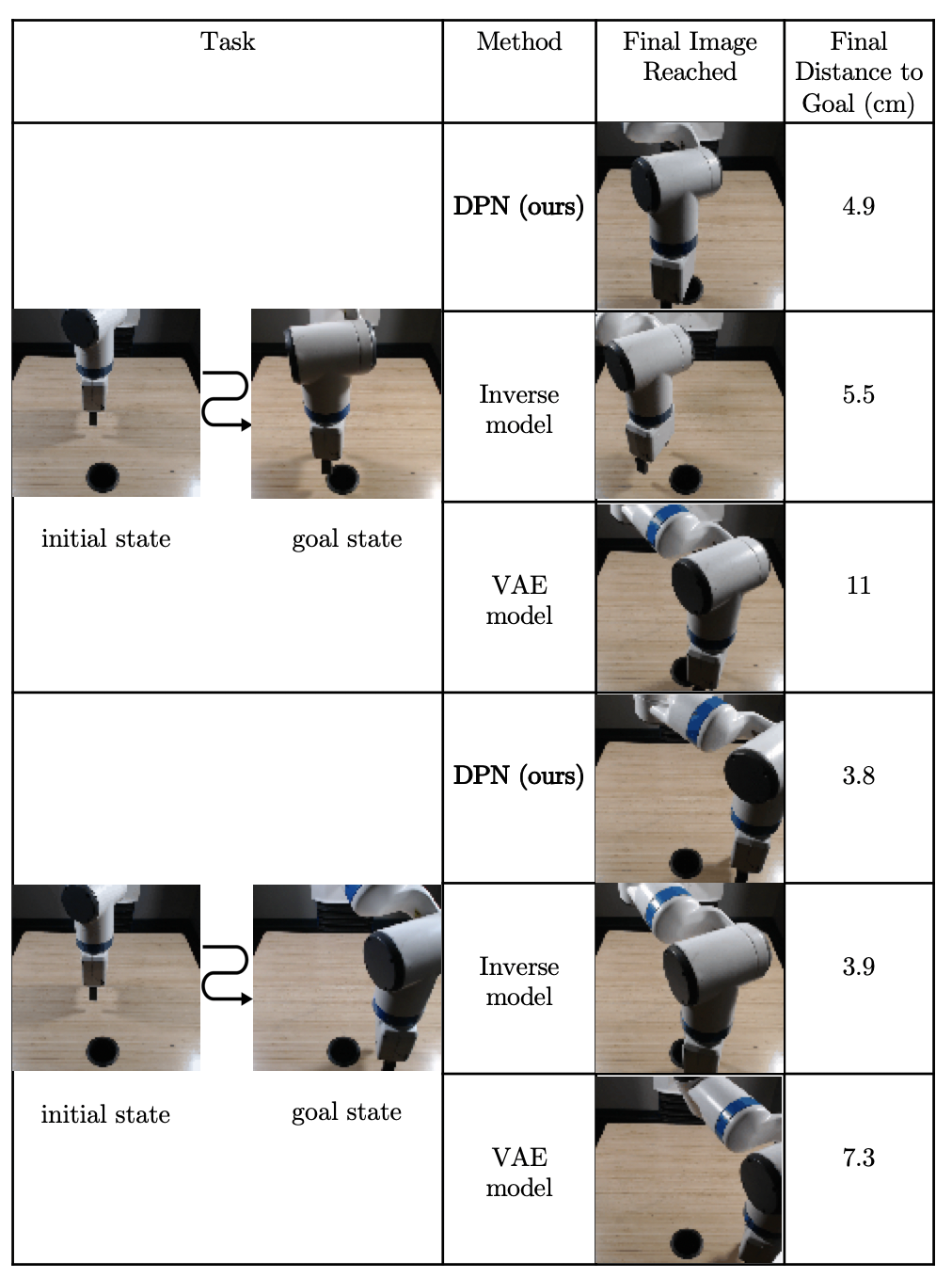}
    \includegraphics[width=0.85\linewidth]{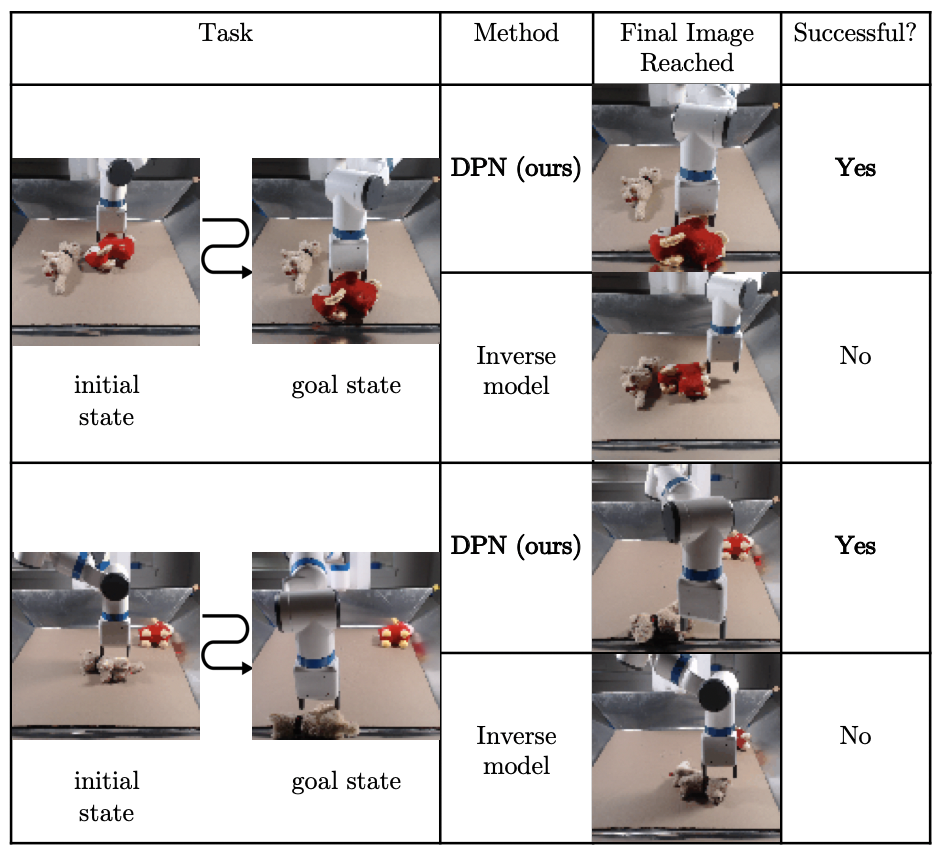}
    \vspace{-0.3cm}
    \caption{Results for the real world reaching and pushing tasks. Our approach is able to learn a metric on real images that enables successful autonomous RL for both reaching and object pushing, whereas prior methods do not consistently lead to successful reinforcement learning.
    }
    \vspace{-0.5cm}
    \label{fig:fetch_results}
\end{figure}

For the first task, the goal image consists of the arm hovering above a hole at the front-middle of the table. 
For the second task, the goal image consists of the arm hovering at the top left corner of the table. 
For both tasks, we run reinforcement learning for approximately $300$ episodes, corresponding to around $3$ hours, for each of the DPN, the inverse model, and VAE metrics. To evaluate performance, we roll out the learned policy $3$ times at the checkpoint that achieved the highest training reward, according to the learned metric, and compute the average true distances to goal position at each final timestep (see Figure~\ref{fig:film_strip}). As shown in Figure~\ref{fig:fetch_results}, we find that the policies that use our DPN metric and the inverse metric performs fairly similarly, while the agent that used the VAE metric performs considerably worse.

Across these two tasks, the VAE model may have performed worse due to slight changes in lighting and camera angle, as previously discussed. Furthermore, as discussed in the simulated reaching example, both DPN and the inverse model likely performed similarly because they both focused on the parts of the videos that correlate most heavily with the actions, i.e. the location of the arm.



\prg{Robot Pushing.}
Seeing that DPN and the inverse model performed similarly on the reaching task, we construct a harder object pushing task that requires the two metrics to pay attention to smaller features in the environment and to take longer-horizon interactions into account.

For this task, the robot has to maneuver a given object from an initial position to a goal position on the table. This task is complicated by a distractor object next to the actual object. We choose two stuffed animals as our given objects and use an aluminum hopper to prevent them from falling off the table during data collection and RL.
Data collection occurs in the same manner as in the reaching experiment. In total, we collect $5000$ episodes, taking approximately one day on one robot.

Since our simulated pushing results indicated that the VAE metric performed poorly, we only compare DPN with the inverse model for this experiment. Again, we trained an agent with each of the DPN and inverse metrics for approximately $400$ episodes and roll the highest-reward policy out $3$ times. As shown in Figure~\ref{fig:fetch_results}, the agent trained with DPN is able to successfully push the stuffed animal to the goal position, while the agent trained with the inverse model is at best only able to push it to the middle of the table. The two metric curves shown in Figure~\ref{fig:fetch_push_metric} give some intuition as to why this is the case. When we plot the metrics for the policy learned w.r.t. the DPN metric, we see that both metrics correctly decrease, although the DPN metric is better at recognizing the similarity of later images, and is thus smaller. Meanwhile, when we plot them for the policy learned w.r.t. the inverse model, we see that the inverse metric incorrectly associates a small latent distance with the earlier images and thus rewards the RL agent for doing nothing, making it difficult for the RL agent to meaningfully move the object. These results seem to match what we saw in the simulated pushing experiment and suggest that DPN does better at distinguishing smaller features that are important for a goal, while the inverse model ignores them and over-prioritizes arm placement.

\section{Discussion} 
\label{sec:conclusion}

\prg{Summary.}
In this paper, we presented an approach for unsupervised learning of a control-centric metric space on images that allows a robot to evaluate its progress towards a specific goal. Our approach proposes more effective and autonomous reinforcement learning while only having access to the goal image by leveraging the learned metric. We then evaluated our method on simulated and real-world tasks, including reaching, pushing, and rope manipulation. Our results suggest that the DPN metric enables RL to perform well on these robotics tasks while converging faster than state-of-the-art techniques for unsupervised goal representations.  

\prg{Limitations and Future Work.}
While we are able to show very good performance for an interesting set of robotics tasks, our method is limited to goal-reaching tasks, which leaves out a number of other interesting RL tasks. Despite this limitation, we believe learning a control-centric metric using our approach may be applicable to a wider range of settings by applying the cost function towards tracking an entire trajectory rather than simply a final state. We are excited to take our approach beyond the current tasks and consider learning control-theoretic metrics in this wider range of settings. 


Beyond vanilla RL, which we study in this paper, a number of other methods rely heavily on effective distance metrics in state space, which have so far limited their application to non-vision domains due to the lack of suitable metrics.
This includes goal relabeling for multi-goal RL~\cite{andrychowicz2017hindsight,pong2018temporal}, planning with learned models~\cite{deepmpc,nagabandi2018neural}, and automatic curriculum generation~\cite{sukhbaatar2017intrinsic}. In future work, we hope to explore the use of our metric in combination with these methods.

\section*{Acknowledgments}
We thank Aravind Srinivas and Sergey Levine for helpful discussions. This work has been partially supported by \url{JD.com} American Technologies Corporation ("JD") under the SAIL-JD AI Research Initiative. This article solely reflects the opinions and conclusions of its authors and not JD or any entity associated with JD.com.


\bibliographystyle{plainnat}
\bibliography{references}

\newpage
\appendix
In this appendix, we summarize our DPN approach in Algorithm~\ref{alg:dpn}. We also provide additional details about our implementation details and experimental setup. 

\subsection{Architecture and Hyperparameters}
\label{app:details}
For all approaches except the $\ell_2$ distance in pixel space, we use a $4$-layer convolutional neural networks with $64$ $5\times 5$ filters each layer and spatial soft-argmax~\cite{dsae} after the last convolution layer to represent the encoder $\enc(\cdot ; \encparams)$, which encodes images into latent space. For DPN, we represent the latent dynamics $\dyn(\x_t, \hat{\action}_t; \dynparams)$ as $2$-layer fully-connected neural networks with $128$ hidden units. The inference network $q(\z_{t} | \action_{t}; \phi)$ is modeled as a $2$-layer fully-connected neural network with hidden units $16$ where the last layer has two heads that output the mean $\mu_\phi(\action_t)$ and the standard deviation $\sigma_\phi(\action_t)$.  The action decoder $\dec(\action_t | \z'_t; \decparams)$ is also a $2$-layer neural network with $16$ hidden units. We use $\beta = 0.5$ as the KL constraint value in $\loss_{\text{DPN}}$. We use $n_p = 20$ as the number of gradient descent steps that update $\z'_{t:t+T}$ and the learned size $\alpha_i$'s are initialized with $0.05$. For the inverse model, we represent the latent multi-step inverse model as a recurrent neural network with $128$ units. For VAE, while the architecture of the encoder is the same across all methods as mentioned above, the decoder consists of $4$ deconvolutional layers with $64$ $5\times 5$ filters except that the last layer has $3$ $5\times 5$ filters in order to reconstruct the image. All three approaches are trained with Adam optimizer~\cite{adam} with learning rate $0.0005$.

\begin{algorithm}[t]
    \caption{Distributional Planning Networks}
    \label{alg:dpn}
\begin{algorithmic}
\REQUIRE random dataset $\{(\obs_{t:t+T+1}, \action_{t:t+T})\}$
\REQUIRE KL constraint value $\beta$, outer step size $\gamma$ 
\STATE Initialize $\alpha_{0:n_p-1}$
\STATE Define the prior $p(\z_{t:t+T})=\mathcal{N}(\mathbf{0},I)$
\WHILE{training}
    \STATE Sample a batch of random data $\obs_t$, $\obs_{t+T+1}$, $\action_{t:t+T}$
    \STATE Sample latent actions $\z_{t:t+T} \sim q_\phi(\z_{t:t+T}| \action_{t:t+T})$
    \STATE Initialize $\z'^{(0)}_{t:T} = \z_{t:t+T}$
    \FOR{$i=0,1,..., n_p-1$}
        \STATE Encode $\x_t = \enc(\obs_t; \encparams)$, $\x_{t+T+1} = \enc(\obs_{t+T+1}; \encparams)$
        \STATE Set $\hat{\x}^{(i)}_t = \x_t$
        \FOR{$j=0, 1, \dots, T$}
            \STATE $\hat{\x}^{(i)}_{t+j+1} = \dyn(\hat{\x}^{(i)}_{t+j}, \z'^{(i)}_{t+j}; \dynparams)$
        \ENDFOR
        \STATE Compute $\loss^{(i)}_{plan} = \loss_\sigma(\hat{\x}^{(i)}_{t+T+1}, \x_{t+T+1})$
        \STATE Update $\z'^{(i+1)}_{t:t+T} = \z'^{(i)}_{t:t+T} - \alpha_i\nabla_{\z'^{(i)}_{t:t+T}}\loss^{i}_{plan}$
    \ENDFOR
    \STATE Compute $\hat{\action}_{t:t+T} = \dec(\z'^{(n_p-1)}_{t:t+T}; \decparams)$
    \STATE Compute $\loss_{\text{DPN}} = \log p_\theta (\action_{t:t+T} | \obs_t, \obs_{t+T+1}) + \beta D_\text{KL}( q_\phi(\z_{t:t+T}| \action_{t:t+T}) ~|| ~ p(\z_{t:t+T}))$
    \STATE Compute $\nabla_\theta\loss_{\text{DPN}}$ and $\nabla_\phi\loss_{\text{DPN}}$
    \STATE Update $\theta \leftarrow \theta - \gamma\nabla_\theta\loss_{\text{DPN}}$
    \STATE Update $\phi \leftarrow \phi - \gamma\nabla_\phi\loss_{\text{DPN}}$
\ENDWHILE
    \STATE Return $\theta, \phi$
\end{algorithmic}
\end{algorithm}

\subsection{Simulated Reaching}

We collect $3000$ videos of the unlabeled interaction by sampling torques uniformly from $[-5, 5]$, where each video consists of $10$ images. Along with each video, we also store a sequence of torques as actions and robot joint angles. For training all methods, we use a training set that contains $28500$ videos and a validation set that has $1500$ videos.

The length of each RL step is $1000$ and the maximum path length is $50$.

\subsection{Simulated Rope Manipulation}

For collecting $20000$ 10-frame random videos, we randomly sample positions of gripper at each time step. We use $19000$ videos for training and the remaining $1000$ for validation.

 We set the maximum path length to be $5$ and the length of one RL step to be $500$.

 \subsection{Simulated Pushing}
 
 At each time step of the random data collection, we apply a random end-effector velocity sampled uniformly from $[-20, 20]$. In this way, we collect $3000$ videos, where $2850$ videos are used for training and $150$ are used for validation.
 
 Each RL step consists of $2000$ timesteps and the maximum path length is $100$ timesteps.

 \subsection{Robot Setup}
 The initial state of the robot is with its arm above the middle of a table. The arm is constrained to move in a square region that is approximately $0.16$m$^2$ in area. In both tasks, we collect $100 \times 100$ RGB images from a front facing camera alongside joint information, and use continuous end effector velocities as actions, which are normalized to be between $[-1, 1]$. Side lights are used to enable nighttime data collection.
 
 \subsection{Robot Reaching}
 
 During unlabeled interation data collection, each episode consists of $20$ timesteps starting from a fixed initial state. At each timestep, we uniformly sample an action to apply to the arm and keep the arm within bounds by applying an inward end effector velocity at table edges.
 
 We use the learned metrics for reinforcement learning of two reaching tasks, where the only reward exposed to the reinforcement learning agent is derived from the latent metric. Because we have access to the true end effector position, we use $\ell_2$ distance between the final position and goal position to evaluate final performance. This distance is not provided to the robot.
 
 In order to make the first task more challenging, we limit the policy to normalized actions between -0.1 and 0.1. This makes it harder for the arm to reach the goal if it initially chooses an incorrect direction of motion. This task, in turn, is designed to see how well each metric works at the start of an episode, where the arm was farther from its goal.
 
 Meanwhile, in the second task, we leave the policy's actions unscaled, making it more likely that the arm will violate bounds and be pushed back inward by the correcting inward velocity once it comes close to an edge. Therefore, this task is meant to show how well each metric performs at the later timesteps of a policy, where the arm is closer to its goal.

 The length of each RL step is $20$ and the maximum path length is $20$.
 
 \subsection{Robot Pushing}
 
 Each episode for random data collection consists of 15 timesteps. In order to vary the initial positions of the objects, we use a mix of scripted shuffling methods and manual rearrangement when objects got stuck in a corner, leading to data collection that is nearly entirely autonomous
 
 For the reinforcement learning task, the arm had to learn to push the either the red or the tan stuffed animal from the middle of the table to a point at the front of the table depending on the goal image (see Fig~\ref{fig:fetch_results}).
 
 The length of each RL step is $20$ and the maximum path length is $20$.

\end{document}